\documentclass{bmvc2k}

\title{Weakly Supervised Generative Network for Multiple 3D Human Pose Hypotheses}

\addauthor{Chen Li}{lic@comp.nus.edu.sg}{1}
\addauthor{Gim Hee Lee}{gimhee.lee@comp.nus.edu.sg}{2}

\addinstitution{
Department of Computer Science,\\
School of Computing,\\
National University of Singapore
}

\runninghead{LI ET AL.}{WEAKLY SUPERVISED GENERATIVE NETWORK FOR 3D  POSE HYPOTHESES}

\def\eg{\emph{e.g}\bmvaOneDot}

\def\etal{\emph{et al}\bmvaOneDot}
\def\ie{\emph{i.e}\bmvaOneDot} 
\def\wrt{w.r.t\bmvaOneDot}

\usepackage{graphicx}
\usepackage{amsmath}
\usepackage{amssymb}
\usepackage{colortbl}
\usepackage{arydshln}
\usepackage{textcomp}
\usepackage{commath}
\usepackage{bbm}
\usepackage{booktabs}
\usepackage{sidecap}
\usepackage[caption=false]{subfig}
\usepackage{wrapfig}

\begin{document}

\maketitle

\begin{abstract}
3D human pose estimation from a single image is an inverse problem 
due to the inherent ambiguity of the missing depth. 
Several previous works addressed the inverse problem by generating multiple hypotheses. However, these works are strongly supervised and require ground truth 2D-to-3D correspondences which can be difficult to obtain. In this paper, we propose a weakly supervised deep generative network to address the inverse problem and circumvent the need for ground truth 2D-to-3D correspondences. To this end, we design our network to model a proposal distribution which we use to approximate the unknown multi-modal target posterior distribution. We achieve the approximation by minimizing the KL divergence between the proposal and target distributions, and this leads to a 2D reprojection error and a prior loss term that can be weakly supervised. Furthermore, we determine the most probable solution as the conditional mode of the samples using the mean-shift algorithm.
We evaluate our method on three benchmark datasets -- Human3.6M, MPII and MPI-INF-3DHP.
Experimental results show that 
our approach is capable of generating multiple feasible hypotheses and achieves state-of-the-art results compared to existing weakly supervised approaches. Our source code is available at: \url{https://github.com/chaneyddtt/weakly-supervised-3d-pose-generator}.
\end{abstract}

\section{Introduction}
\label{sec:intro}
3D human pose estimation from a monocular image refers to the task of recovering 3D human pose from a 2D image of the person. This task is extensively studied in the computer vision community due to its potentially useful applications in surveillance, healthcare, movie productions, robotics, etc. Most existing works for the task of 3D human pose estimation from a monocular image assume a uni-modal posterior distribution where only a single solution can exist. On the contrary, following the arguments by \cite{Li_2019_CVPR, jahangiri2017generating}, we reason that 3D human pose estimation from a monocular image is actually an inverse problem with the possibility of multiple feasible solutions due to the inherent ambiguity of the missing depth. Enforcing a uni-modal posterior distribution on the models can lead to overfitting that gives undesirable performance. 

To the best of our knowledge, the only existing works that addressed the inverse problem of 3D human pose estimation from a monocular image are Jahangiri and Yullie \cite{jahangiri2017generating}, and Li and Lee \cite{Li_2019_CVPR}. More specifically, \cite{jahangiri2017generating} uses optimization based method that generates multiple hypotheses for the inverse problem. Despite the ability to generate multiple hypotheses, the method shows unsatisfactory performance compared to existing deep learning approaches that produce only a single solution. \cite{Li_2019_CVPR} is the first and currently the only deep learning approach that generates multiple hypotheses for the inverse problem of 3D human pose estimation. It uses a mixture density network to model the posterior with a multi-modal  mixture-of-Gaussian distribution. Although this approach outperforms other state-of-the-art deep learning single solution approaches, it is supervised that requires a huge amount of ground truth 2D-to-3D correspondences that are often difficult to obtain. To circumvent the need for ground truth data, an increasing number of weakly supervised \cite{Wandt2019RepNet, tung2017adversarial, drover2018can, kanazawa2018end} and unsupervised \cite{rhodin2018unsupervised}  deep learning approaches are proposed in the recent years. However, these approaches are still based on a uni-modal posterior assumption that gives a single solution to the inverse problem of 3D human pose estimation from a monocular image. 

\begin{SCfigure}
  \includegraphics[width=0.6\textwidth]{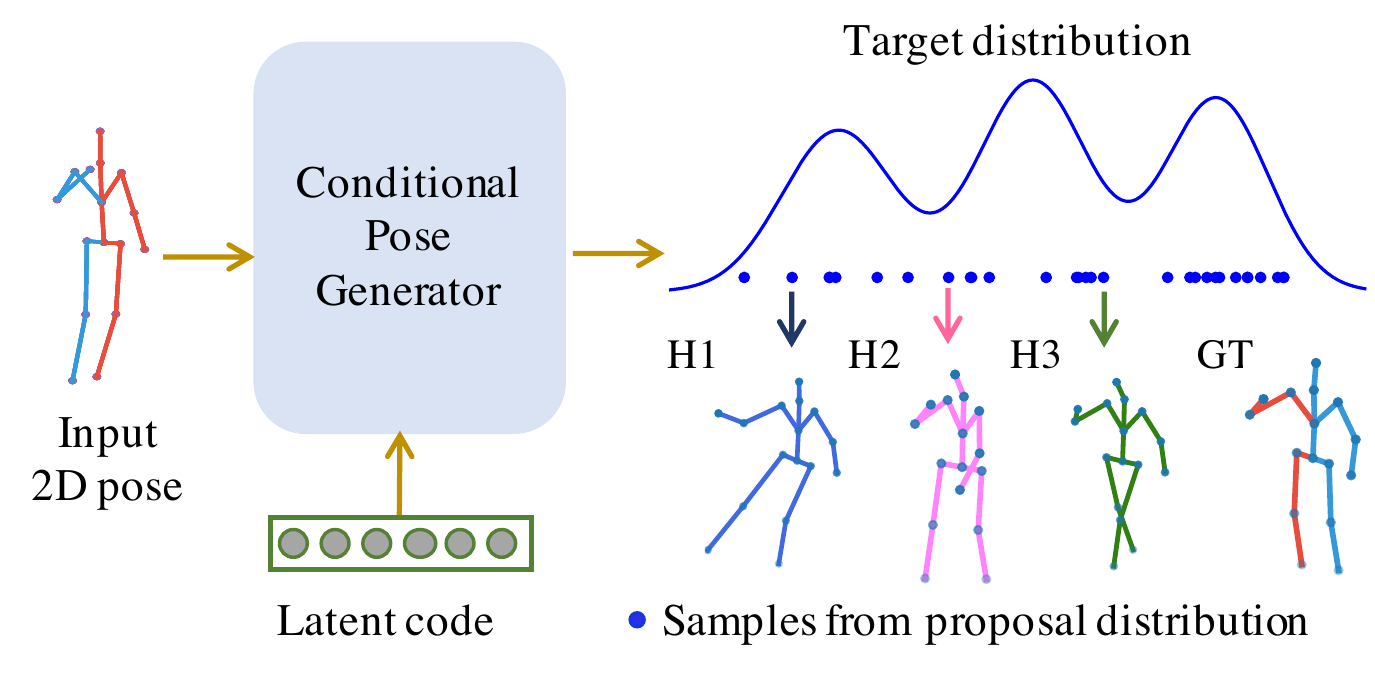}
    \label{fig:teaser}
      \caption{Our deep generative network is conditioned on a input 2D pose. Latent codes are drawn from a normal distribution to generate samples of 3D pose hypotheses that correspond to the target multi-modal posterior distribution.}
  \vspace{-3mm}
\end{SCfigure}

In this paper, we propose a weakly supervised deep generative network to address the inverse problem of 3D human pose estimation. To this end, we design a deep generative network to model a proposal distribution which we use to approximate the unknown multi-modal posterior distribution. Figure \ref{fig:teaser} shows an illustration of our approach. We achieve the approximation by minimizing the KL divergence between our proposal distribution and the target posterior distribution. This leads to a loss function that minimizes the expectation of a 2D reprojection error and a prior term over the samples drawn from the proposal distribution. The 2D reprojection error ensures that samples of the 3D human pose drawn from our deep generative network reproject closely to the 2D pose observed in the image. We use a discriminator based on the maximum mean discrepancy (MMD) \cite{li2017mmd, binkowski2018demystifying} as the prior term to encourage the generated 3D human pose to be ``human-like". Furthermore, we prevent the mode collapse problem of our generative network by introducing two additional losses \cite{yang2019diversity, zhu2017toward}  into the prior term to encourage diversity in the generated 3D human poses.


Given an input 2D human pose during inference, we draw samples from the posterior distribution by generating multiple 3D human poses from our generative network. We determine the most probable solution as the conditional mode of the samples using the mean-shift algorithm. We further propose a time-efficient variant to approximate the conditional mode. Specifically, we approximate the conditional mode as the output of our generative network from an all-zero latent code input.
Experimental results show that our approach can achieve superior performance compared to state-of-the-art weakly supervised approaches on the Human3.6M dataset \cite{ionescu2013human3}. We also test on the MPII \cite{andriluka20142d} and the MPI-INF-3DHP datasets \cite{mehta2017monocular} to show the generalization capacity. Our contributions are summarized as:
(1) We propose a weakly supervised deep generative network to generate multiple hypotheses for the inverse problem of 3D human pose estimation. 
(2) We prevent mode collapse of our network by introducing additional losses to encourage diversity of the generated hypotheses.
(3) We achieve state-of-the-art results compared to other weakly supervised approaches.
\vspace{-2mm}

\section{Related work}
Existing 3D pose estimation approaches can be divided into three categories: Fully and weakly supervised approaches based on a uni-modal posterior, and fully supervised approaches based on a mixture-of-Gaussians distribution. 

Most existing works are fully supervised, which train their models either in an end-to-end \cite{lee2018propagating, mehta2017monocular, pavlakos2017coarse, sun2018integral,zhou2017towards} or a two-stage manner \cite{bogo2016keep, martinez2017simple, rayat2018exploiting,moreno20173d, yasin2016dual}. Pavlakos \etal \cite{pavlakos2017coarse} use a volumetric representation for the 3D space and train a deep network to estimate the probability that a joint is located at each voxel. Because of the high dimension of the output, a coarse-to-fine strategy is adopted to finetune the estimation iteratively. To improve the generalization capacity, Zhou \etal \cite{zhou2017towards} proposes a transfer learning approach such that the network can be trained with both outdoor and indoor images. For the two-stage approaches, Martines \etal \cite{martinez2017simple}  use a simple deep neural network to estimate 3D pose from 2D joint detections. Despite the impressive results, these approaches require ground truth 2D-to-3D labels, which are tedious to collect especially for outdoor environments.

More recently, several works begin to focus on weakly supervised \cite{Wandt2019RepNet, tung2017adversarial, drover2018can}, unsupervised \cite{rhodin2018unsupervised} and self-supervised learning \cite{chen2019unsupervised}. Wandt \etal \cite{Wandt2019RepNet} weakly supervise their network with only 2D ground truth labels by projecting the estimated 3D pose into 2D space. A critic network is then used to enforce the estimated poses to be realistic.  Chen \etal \cite{chen2019unsupervised} propose a self-supervised learning framework that lifts the 2D input to 3D pose, projects the 3D pose after a random transform, lifts the projection to 3D, undo the random transform and then projects back onto the 2D image. A self-consistency constraint and a 2D pose discriminator is applied on the original and final 2D poses to enable the lifting network to estimate valid 3D poses. The discriminators applied in both approaches play a key role to enforce valid estimations.

All of the above mentioned approaches assume a uni-modal posterior distribution, and only estimate one 3D pose for each 2D input. Two recent works \cite{Li_2019_CVPR, jahangiri2017generating} explore a new line of research in generating multiple hypotheses for 3D human pose estimation. They argue that 3D pose estimation from 2D observations is an inverse problem where multiple solutions exist. To generate the multiple solutions, Jahangiri and Yullie \cite{jahangiri2017generating} learn an occupancy matrix to represent the plausible angular regions for each joint, and then generate multiple hypotheses by sampling from the occupancy matrix. Li and Lee \cite{Li_2019_CVPR} use a mixture density network (MDN) to learn the multi-modal posterior distribution and take the conditional mean values of the mixture-of-Gaussian distribution as the hypotheses. Although \cite{Li_2019_CVPR} achieves promising results, the method is strongly supervised and require ground truth 2D-to-3D correspondences for training. In contrast to \cite{Li_2019_CVPR, jahangiri2017generating}, we propose a weakly supervised generative model to generate multiple 3D pose hypotheses.

\vspace{-3mm}

\section{Our Method}
\begin{figure*}
\begin{center}
\includegraphics[width=0.98\linewidth]{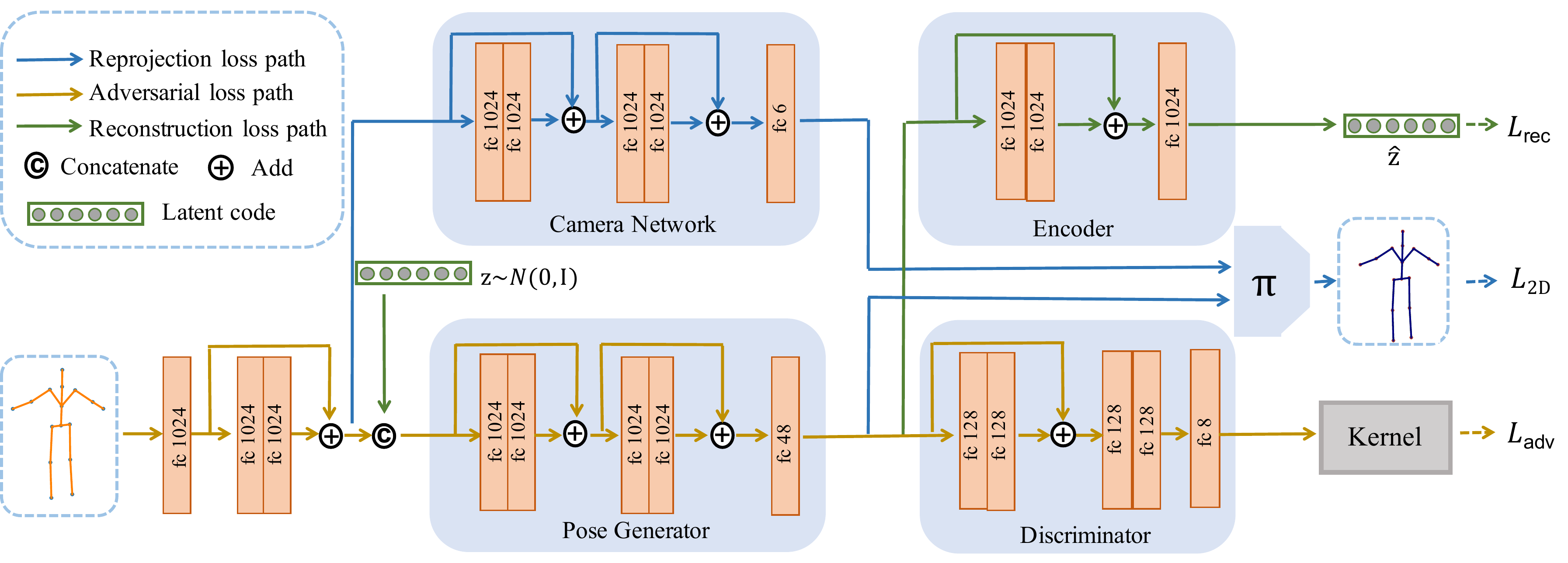}

\end{center}
  \vspace{-4mm}
   \caption{Our deep generative network to generate multiple 3D human pose hypotheses. 
   }
 
\label{fig:OurNetwork}
\vspace{-3mm}
\end{figure*}

We propose a weakly supervised approach to generate multiple hypotheses from a given 2D human pose input. Let $\mathbf{x} \in \mathbb{R}^{2C} $ denotes the 2D joint detection, where $C$ is number of joints in a skeleton. We generate multiple 3D pose hypotheses $\mathbf{y} \in \mathbb{R}^{3C}$ for each 2D pose input, where all the hypotheses reproject close to the 2D pose input. The true posterior $P(\mathbf{y}\mid\mathbf{x})$ is a multi-modal distribution because of the depth ambiguity and occluded joints. We design a deep generative network as a proposal distribution $Q(\mathbf{y}\mid\mathbf{x})$ to approximate the unknown target posterior distribution $P(\mathbf{y}\mid\mathbf{x})$. 
Figure~\ref{fig:OurNetwork} shows the deep generative network that we designed as the proposal distribution $Q(\mathbf{y}\mid\mathbf{x})$. It consists of four main components: (1) a pose generator network that generates a 3D pose hypothesis $\mathbf{y}$ from on an input 2D pose $\mathbf{x}$ and latent code $\mathbf{z} \sim \mathcal{N}(0,\mathbf{I})$; (2) a camera network that estimates the camera matrix $\mathbf{M}\in \mathbb{R}^{2 \times 3}$ to project the generated 3D pose hypotheses into the 2D space; (3) a discriminator as the prior $P(\mathbf{y})$ of the generated 3D pose; and (4) an encoder as a second prior to prevent the model collapse of our generative model.

\subsection{Conditional Pose Generator}
Our goal is to train the model $Q(\mathbf{y}\mid\mathbf{x})$ to generate samples of 3D pose hypotheses from the unknown target posterior distribution $P(\mathbf{y}\mid\mathbf{x})$. To this end, we minimize the KL divergence between the proposal $Q(\mathbf{y}\mid\mathbf{x})$ and the target posterior $P(\mathbf{y}\mid\mathbf{x})$ distributions:
\begin{align}
  \label{eq:kl_divergence}
  \mathcal{L} = KL[Q(\mathbf{y}\mid \mathbf{x}) \| P\mathbb(\mathbf{y}\mid \mathbf{x})] + H(Q(\mathbf{y}\mid \mathbf{x})).
\end{align}
Following \cite{chen2016infogan}, we also minimize the entropy of the proposal distribution so that the output 3D pose $\mathbf{y}$ learns enough information from the 2D input $\mathbf{x}$. 
According to the definition of KL divergence and entropy, the objective function is evaluated as:
\begin{align}
  \label{eq:derivation}
  \mathcal{L} = - \sum_y Q(\mathbf{y}\mid\mathbf{x}) \log \frac{P(\mathbf{y} \mid \mathbf{x})}{Q(\mathbf{y} \mid \mathbf{x})}
   - \sum_y Q(\mathbf{y} \mid \mathbf{x}) \log  Q(\mathbf{y} \mid \mathbf{x}) 
  =  - \sum_y \{Q(\mathbf{y} \mid \mathbf{x}) \log P(\mathbf{y} \mid \mathbf{x})\} .
\end{align}
We get our final objective function by applying the Bayes rule on $P(\mathbf{y} \mid \mathbf{x}) = \frac{P(\mathbf{x} \mid \mathbf{y}) P(\mathbf{y})}{P(\mathbf{x})}$:  
\begin{align}
    \label{eq:objective}
    \mathcal{L}
    = -\mathop{\mathbb{E}}_{y \sim Q(\mathbf{y} \mid \mathbf{x})}\{\log P(\mathbf{x} \mid \mathbf{y}) + \log P(\mathbf{y})\}.
\end{align}
The objective function consists of a likelihood term $P(\mathbf{x} \mid \mathbf{y})$ and a prior $P(\mathbf{y})$ term after we drop the constant term $\log P(\mathbf{x})$. We represent the likelihood term $P(\mathbf{x} \mid \mathbf{y})$ by a Laplace distribution:
\begin{align}
    \label{eq:likelihood}
     P(\mathbf{x} \mid \mathbf{y}) = \frac{1}{2b}\exp{-\frac{|{\pi(\mathbf{y}) -\mathbf{x}}|}{b}},
\end{align}
where $b$ is the scale parameter, $\pi(\mathbf{y})$ and $\mathbf{x}$ are the 2D reprojection of the generated 3D pose and the input 2D pose, respectively. Note that the Gaussian or the Laplacian distribution can be used for the likelihood term, and we chose the Laplacian distribution due to its robustness to noisy and outlier 2D joint inputs. Inspired by \cite{Wandt2019RepNet, habibie2019wild}, we estimate a camera matrix $\mathbf{M}\in \mathbb{R}^{2 \times 3}$ from the 2D observation by using a camera network. The generated 3D pose $\mathbf{y}$ and camera matrix $\mathbf{M}$ are fed into a reprojection module to get the 2D reprojection. Under a weak perspective camera assumption, the 2D reprojection of the generated pose $\mathbf{y}$ is given by 
$\pi(\mathbf{y}) = \mathbf{M}\mathbf{y}$.
Maximizing the log-likelihood term is equivalent to minimizing the reprojection error, which results in our 2D loss: 
$\mathcal{L}_{\text{2D}} = | {\mathbf{M}\mathbf{y} -\mathbf{x}}|$.

The prior term $P(\mathbf{y})$ represents the prior knowledge of real 3D poses, \eg bone length, joint angle limit and symmetric information, and we use the discriminator from the MMD GAN \cite{li2017mmd, binkowski2018demystifying} to learn the prior knowledge from a set of 3D poses. Note that it is not necessary for this set of 3D poses to be the ground truth labels of the respective input 2D poses. 
The input to the discriminator is a concatenation of the 3D pose and the corresponding KCS matrix \cite{Wandt2019RepNet}. 
As shown in Figure~\ref{fig:OurNetwork}, our pose generator has similar structure to a conditional GAN. The generator generates pose hypotheses from latent code $\mathbf{z} \sim \mathcal{N}(0, \mathbf{I})$ conditioned on the input 2D pose $\mathbf{x}$, while the discriminator try to distinguish the generated poses from real poses. Consequently, a sample $\mathbf{y}$ is drawn from the proposal distribution in Equation~\eqref{eq:objective} as:
\begin{align}
    \label{sampling}
    \mathbf{y} \leftarrow Q(\mathbf{y} \mid \mathbf{x}, \mathbf{z} \sim \mathcal{N}(0, \mathbf{I})).
\end{align}

\subsection{Diverse Pose Hypotheses}
Minimization of the KL divergence between the proposal distribution and the target posterior distribution may result in the generator learning only a subset of the target posterior distribution.  This phenomenon
known as the mode collapse problem \cite{salimans2016improved,arjovsky2017principled} is widely discussed in the GAN literature. This problem manifests itself in the generator generating the same poses for different input latent codes conditioned on the same input 2D pose. To circumvent this problem, we add a second prior with a regularizer \cite{yang2019diversity} to explicitly encourage diversity and an encoder to reconstruct the input noise \cite{zhu2017toward}.

Let $G(\mathbf{x}, \mathbf{z}_1)$ and $G(\mathbf{x}, \mathbf{z}_2)$ denote the output of the generator given 2D observation $\mathbf{x}$, latent codes $\mathbf{z}_1$ and $\mathbf{z}_2$ sampling from $\mathcal{N}(0, \mathbf{I})$. We encourage the generator to generate diverse hypotheses by maximizing the objective:
\begin{align}
\label{eq:regularization}
    \mathcal{L}_\text{reg} = \mathbb{E}_{\mathbf{z}_1, \mathbf{z}_2} [\min (\frac{|G(\mathbf{x}, \mathbf{z}_1) - G(\mathbf{x}, \mathbf{z}_2)|}{|\mathbf{z}_1-\mathbf{z}_2|}, \tau)],
\end{align}
where $\tau$ is a constant to ensure numerical stability. The regularizer forces the generator to generate diverse poses depending on the distance between the input latent codes. 

To further prevent the mode collapse, we also introduce another encoder $E$ to reconstruct the input latent code \cite{zhu2017toward}:
\begin{align}
\label{eq:reconstruction}
    \mathcal{L}_\text{rec} = \mathbb{E}_{\mathbf{z} \sim \mathcal{N}(0, 1)} {|\mathbf{z} - E(G(\mathbf{x}, \mathbf{z}))|}.
\end{align}
The reconstruction loss encourages the connection between the output 3D pose and input latent code to be invertible, such that it helps prevent the many-to-one mapping problem in mode collapse. Intuitively, if $G(\mathbf{x}, \mathbf{z}_1)$ and $G(\mathbf{x}, \mathbf{z}_2)$ are the same when $\mathbf{z}_1 \neq \mathbf{z}_2 $, we can never recover $\mathbf{z}_1 $ or $\mathbf{z}_2$ because the inputs to the encoder $E$ are the same.

\subsection{Optimization}
Inspired by the MMD GAN \cite{li2017mmd, binkowski2018demystifying}, we use the kernel maximum mean discrepancy to distinguish the generated and real data distributions. The unbiased estimator of the squared MMD is given by:
\begin{align}
    MMD_u^2(P,Q) = \frac{1}{m(m-1)}\sum_{i \neq j}^m k(\mathbf{y}_i, \mathbf{y}_j) + \frac{1}{n(n-1)}\sum_{i \neq j}^m k(\tilde{\mathbf{y}}_i, \tilde{\mathbf{y}}_j) - \frac{2}{mn}\sum_{i=1}^m \sum_{j=1}^n k(\mathbf{y}_i, \tilde{\mathbf{y}}_j).
\end{align}
where $\mathbf{y} \sim P(\mathbf{y})$ and $\tilde {\mathbf{y}} \sim Q(\mathbf{y} \mid \mathbf{x})$ represent samples from the real and generated distributions respectively. We adopt a mixed kernel consisting of the rational quadratic (RQ) kernel and the dot kernel: $k^{rq*} = k^{rq} + k^{dot}$ following \cite{binkowski2018demystifying}, where
\begin{align}
    k_\alpha^{rq}(x_1,x_2) = (1 + \frac{\norm{x_1-x_2}^2}{2\alpha})^{-\alpha} ,  k^{dot}(x_1, x_2) = \langle x_1,x_2 \rangle.
\end{align}

The pose generator tries to fool the discriminator by generating realistic poses, hence it minimizes a adversarial loss given by: $\mathcal{L}_\text{adv} = MMD_u^2(P, Q)$. At the same time, the pose generated from the same 2D input should be diverse and also keep consistent with the 2D input. Finally, the full objective function of the generator is expressed as:
\begin{align}
    \label{loss_generator}
    \mathcal{L}_G = \mathcal{L}_\text{adv} + \lambda_{\text{2D}}\mathcal{L}_{\text{2D}} - \lambda_{\text{reg}}\mathcal{L}_{\text{reg}} + \lambda_{\text{rec}}\mathcal{L}_{\text{rec}},
\end{align}
where $\lambda_{\text{2D}}$,  $\lambda_{\text{reg}}$ and $\lambda_{\text{rec}}$ represent the weights of the corresponding losses. On the other hand, the discriminator tries to distinguish the real and fake distributions by minimizing $\mathcal{L}_D = - \mathcal{L}_\text{adv} + \lambda_{\text{gp}}\mathcal{L}_\text{gp}$. The gradient penalty $\mathcal{L}_\text{gp}$ term \cite{arjovsky2017wasserstein} is added to enforce the Lipschitz constraint. The camera estimation network also optimizes a camera loss  \cite{Wandt2019RepNet}
such that it fulfils the weak perspective camera constraint.

\subsection{Best Pose Selection}
\label{pose_selection}
After training, the generator can generate 3D pose hypotheses for the same 2D input by sampling latent code $\mathbf{z}$ from $\mathcal{N}(0, \mathbf{I})$. In practice, we also want to find the most probable 3D pose from the multiple hypotheses, \ie, the best conditional mode of the posterior distribution. 
Let $\mathbf{H} = \{\mathbf{h}_1, \mathbf{h}_2, \mathbf{h}_3, ..., \mathbf{h}_N \}$ be the pose hypotheses generated from $\mathbf{Z} = \{\mathbf{z}_1, \mathbf{z}_2, \mathbf{z}_3, ..., \mathbf{z}_N\}$ conditioned on $\mathbf{x}$, where $N$ is the number of samples. To find the pose with the highest probability, we employ a local mode-finding approach based on mean-shift \cite{comaniciu2002mean} with a Gaussian kernel. The kernel density estimator is given by: 
\begin{align}
    \label{kernel_meanshift}
    \hat{f}(\mathbf{h}) = \frac{1}{Nw^d}\sum_{i=1}^NK(\frac{\mathbf{h} - \mathbf{h}_i}{w}),
\end{align}
where $w$, $d$ and $K$ represent the bandwidth, feature dimension and kernel function, respectively.
However, the mean-shift algorithm is computationally expensive especially when the number of samples is large and might be unsuitable for scenario where efficiency is the priority. We propose an alternative method to improve the efficiency. We directly feed an all-zero code $\mathbf{z}$ into the generator and obtain the final pose. This is similar to the `zero code' used in \cite{zhi2019scenecode, bloesch2018codeslam} to obtain a most likely single view depth. Intuitively, an all-zero code is the most likely code because we sample $\mathbf{z}$ from $\mathcal{N}(0, \mathbf{I})$ during training. 
We will show in the experiments that the zero code can achieve comparable results with the mean-shift algorithm.
\vspace{-4mm}

\section{Experiments}
\paragraph{Implementation Details.} We train our model with ADAM optimizer with an initial learning rate of 0.0001 and decay every epoch with a decay rate of 0.94. The weights for different losses $\lambda_{\text{gp}}$, $\lambda_{\text{2D}}$, $\lambda_{\text{reg}}$ and $\lambda_{\text{rec}}$ are set to 0.1, 10.0, 7.5 and 10.0 respectively. 
 \vspace{-3mm}

\paragraph{Datasets.} We evaluate our approach on three 3D human pose estimation benchmarks: Human3.6M \cite{ionescu2013human3}, MPI-INF-3DHP \cite{mehta2017monocular} and MPII datasets\cite{andriluka20142d}. The human3.6M dataset is the largest and most commonly used dataset for 3D human pose estimation. There are 15 daily activities in total performed by 7 professional actors under 4 camera views. The MPI-INF-3DHP is a recently proposed dataset which includes both indoor and outdoor scenes. The MPII dataset a challenging benchmark for 2D human pose estimation because of the complex background and severe occlusion. We train our model on the Human3.6M dataset and show results on all three datasets
\vspace{-4mm}
\paragraph{Data Preprocessing.}  Following previous work \cite{Wandt2019RepNet}, we align every 3D pose in the Human3.6M dataset to a template by applying a transformation to the 3D pose. The transformation, which includes a scale, rotation and translation, is obtained from procrustes analysis on the hip and shoulder joints. Both 2D and 3D poses are centered at the root joint, and each 2D pose is further normalized by dividing its standard deviation. Following previous work \cite{Li_2019_CVPR, martinez2017simple}, we use the stacked hourglass network \cite{newell2016stacked} trained on both MPII and Human3.6M datasets to estimate 2D poses from images.

\vspace{-4mm}
\paragraph{Evaluation Protocols.} Following the standard protocol for Human3.6M dataset \cite{kocabas2019self}, we use subjects 1, 5, 6, 7 and 8 for training, and evaluation is done on every $64^{th}$ frame of subjects 9 and 11. The evaluation metric is the Mean Per Joint Position Error (MPJPE) measured in millimeters. 
The 3D Percentage of Correct Keypoints (3DPCK) under 150mm radius \cite{mehta2017monocular} is adopted as the metric for the MPI-INF-3DHP dataset.

\begin{table*}[h!]
\centering
\scriptsize
\setlength{\tabcolsep}{0.7pt}
\begin{tabular*}{1\textwidth}{ l c c c c c c c c c c c c c c c c c c } 
 
 \hline
Protocol \#2 & MH & WS & Direct. & Discuss & Eating & Greet & Phone & Photo & Pose & Purch. &  Sitting & SitD. & Smoke  & Wait & WalkD. & Walk & WalkT. & Avg.\\ 
 \hline
 
 Martinez \cite{martinez2017simple} & &   &39.5 &  43.2 & 46.4 & 47.0 &  51.0 & 56.0  & 41.4 & 40.6  & 56.5  &  69.4  & 49.2 &  45.0 &  49.5 & 38.0 &  43.1 & 47.7\\
 
 Zhou \cite{zhou2019hemlets}  & & &29.1 & 34.9 & 29.9  & 32.6 & 31.2 & 32.3 & 27.0 & 33.3 & 37.6 & 45.9 & 32.2 &31.5   & 34.5 & 22.9 & 25.9 &  32.1\\
 
 Li \cite{Li_2019_CVPR}(BH) & \checkmark &  & 35.5 & 39.8  & 41.3  & 42.3  & 46.0  & 48.9 & 36.9  & 37.3 &  51.0
 & 60.6 & 44.9  &  40.2  &  44.1  & 33.1  & 36.9  &  42.6\\
 
  Tung \cite{tung2017adversarial} & & \checkmark &77.6 & 91.4 & 89.9  &88.0  & 107.3 & 110.1 & 75.9 & 107.5 & 124.2 & 137.8  & 102.2 & 90.3 & 78.6  & - & -  & 97.2 \\
  
 Wandt \cite{Wandt2019RepNet} &  &\checkmark  & 53.0 & 58.3 & 59.6 & 66.5 &72.8 &71.0 & 56.7 & 69.6 & 78.3 & 95.2 & 66.6 & 58.5 & 63.2 & 57.5 &49.9 & 65.1 \\
 Drover \cite{drover2018can} &  &\checkmark & 60.2 & 60.7 & 59.2 & 65.1 & 65.5 & 63.8 & 59.4 & 59.4 & 69.1 & 88.0 & 64.8 & 60.8 & 64.9 & 63.9 & 65.2 & 64.6 \\
\hdashline
 Ours (ZC) &\checkmark &\checkmark &  \textbf{42.1} & \textbf{44.7}  & \textbf{45.4}  & \textbf{51.0}  & \textbf{49.3}  & \textbf{51.5} & \textbf{41.2}  & \textbf{46.2} & \textbf{57.5} 
 & \textbf{70.8}  & \textbf{48.7}  & \textbf{44.1}  & \textbf{50.8} & \textbf{42.1} & \textbf{43.7}  & \textbf{48.7} \\
 Ours (MS) &\checkmark &\checkmark & 41.4 & 44.3 & 44.6 & 50.2 & 49.3 & 51.8 & 40.1 & 46.2 & 57.7 & 72.7 & 48.7 & 45.4 & 49.6 & 43.8 & 43.3  & 48.7\\

 Ours (BH) &\checkmark &\checkmark  &38.5 & 41.7  & 39.6  & 45.2  & 45.8  & 46.5 & 37.8  & 42.7 &  52.4
 & 62.9 & 45.3  &  40.9  &  45.3 & 38.6 & 38.4   & 44.3 \\

 Ours (GT+BH) &\checkmark &\checkmark & 26.8 & 31.2 & 26.9 & 33.0 & 31.0 & 36.9 & 28.7 & 31.2 & 36.6 & 46.4 & 30.0 & 30.8 & 31.5 & 24.7 & 27.2  & 31.6\\
\hline
\end{tabular*}
\vspace{-3mm}
\caption{Quantitative results of MPJPE on the Human3.6M dataset under protocol \#2. The best results for weakly supervised methods are marked in bold. (Our results under ZC setting is used for fair comparison.)}
\label{Tab:MPJPE_Results_Human3.6}
\end{table*}
\vspace{-5mm}

\subsection{Quantitative Results on Human3.6M Dataset}
The poses generated by the generator are in the template frame as described in the Data Preprocessing. Consequently, we evaluate the effectiveness of the pose generator by showing the MPJPE under protocol \#2, where a rigid alignment is applied to the estimated pose before comparison with the ground truth. Table \ref{Tab:MPJPE_Results_Human3.6} shows the results of our approach and other state-of-the-art fully and weakly supervised approaches. `MH' represents approaches that generate multiple hypotheses and `WS' represents weakly supervised approaches. We evaluate our approach under both mean-shift (MS) and zero code (ZC) settings, where we generate a single pose with the highest probability \wrt the proposal distribution. Following previous works \cite{Li_2019_CVPR, jahangiri2017generating} that generate multiple hypotheses, we also evaluate our approach under the best hypothesis (BH) setting, where we select the best of ten hypotheses according to the ground truth. As can be seen from Table \ref{Tab:MPJPE_Results_Human3.6}, our approach achieves comparable performance with our supervised counterpart \cite{Li_2019_CVPR}, which also generates multiple hypotheses, and superior results compared to other weakly supervised approaches \cite{tung2017adversarial, Wandt2019RepNet, drover2018can}. The similar performance achieved by MS and ZC demonstrates that the pose with highest probability can be approximated from the zero code. This significantly improves the efficiency because sampling is not needed in ZC setting. Moreover, the performance under ZC (or MS) is close to BH. This shows that the pose hypothesis with the highest probability is close to the ground truth pose among all hypotheses generated by the generator. `GT' represents results when using ground truth 2D joints as input, which indicates that our performance can be further improved when 2D detections are more accurate.

\begin{table*}
\centering
\scriptsize
\setlength{\tabcolsep}{0.7pt}
\begin{tabular*}{0.99\textwidth}{ l c c c c c c c c c c c c c c c c c c } 
 
 \hline
  Protocol \#1  & MH & WS &  Direct. & Discuss & Eating & Greet & Phone & Photo & Pose & Purch. &  Sitting & SitD. & Smoke & Wait & WalkD. & Walk & WalkT. & Avg.\\ 
 \hline

 Martinez \cite{martinez2017simple} & & & 51.8 & 56.2 & 58.1 & 59.0 & 69.5 & 78.4 & 55.2 & 58.1 & 74.0 & 94.6 & 62.3 & 59.1 & 65.1 & 49.5 & 52.4 & 62.9 \\
 
 Sun \cite{sun2018integral} & & &47.5 & 47.7 & 49.5 & 50.2 & 51.4 & 55.8 & 43.8 & 46.4 & 58.9 & 65.7 & 49.4 & 47.8 & 49.0&   38.9 &  43.8 &  49.6 \\
 Zhou \cite{zhou2019hemlets} & & & 34.4 & 42.4 & 36.6 & 42.1 & 38.2 & 39.8 & 34.7 & 40.2 &  45.6 & 60.8 & 39.0 & 42.6 &  42.0 &  29.8 &  31.7 &  39.9 \\
 
 
 
 Li \cite{Li_2019_CVPR}(BH) & \checkmark &  & 43.8 & 48.6 & 49.1  & 49.8 & 57.6 & 61.5  & 45.9 & 48.3 & 62.0 & 73.4 & 54.8 & 50.6 &  56.0 & 43.4 & 45.5 & 52.7 \\
  
 Wandt \cite{Wandt2019RepNet} &  &\checkmark  & 77.5 & 85.2 & 82.7 & 93.8 &93.9 &101.0 & 82.9 & 102.6 & 100.5 & 125.8 & 88.0 & 84.8 & 72.6 & 78.8 &\textbf{79.0} & 89.9 \\

 \hdashline
  Ours (ZC) &\checkmark &\checkmark &  \textbf{67.9} & \textbf{75.5}  & \textbf{71.8}  & \textbf{81.8}  & \textbf{81.4}  & \textbf{93.7} & \textbf{75.2}  & \textbf{81.3} & \textbf{88.8} 
 & \textbf{114.1} & \textbf{75.9}  & \textbf{79.1} & \textbf{83.3} & \textbf{74.3} & \textbf{79.0}    & \textbf{81.1} \\
 
 Ours (MS) &\checkmark &\checkmark &  66.0 & 74.7  & 71.1  & 80.6  & 81.1  & 93.0 & 73.2  & 83.7 &  90.0
 & 117.4 & 75.8  &  79.3  &  82.1  & 74.4 & 77.8  & 80.9 \\
 
  Ours (BH) &\checkmark &\checkmark  & 62.0 & 69.7  & 64.3  & 73.6  & 75.1  & 84.8 & 68.7  & 75.0 &  81.2
 & 104.3 & 70.2  & 72.0  &  75.0  & 67.0 & 69.0   & 73.9 \\
 
 Ours (GT+BH) &\checkmark &\checkmark & 54.8 & 61.9 & 48.6 & 63.6 & 55.8 & 73.7 &59.0 & 61.3 & 62.2 & 85.7 & 52.8 & 60.2 &57.5 & 51.3 & 56.8 & 60.0\\

\hline
\end{tabular*}
\caption{Quantitative results of MPJPE on the Human3.6M under protocol \#1.The best results for weakly supervised methods are marked in bold. (Our results under ZC setting is used for fair comparison.)}
\label{Tab:MPJPE_Results_Human3.6_Camera_Frame}
\vspace{-5mm}
\end{table*}

We evaluate our model under protocol \#1, where the generated poses are transformed into the camera frame with a rotation matrix $\mathbf{R}$ computed from the camera network output $\textbf{M} \in \mathbb{R}^{2 \times 3}$. As shown in Table~\ref{Tab:MPJPE_Results_Human3.6_Camera_Frame}, our approach outperforms state-of-the-art weakly supervised approach \cite{Wandt2019RepNet}.
Note that our approach performs worse than our supervised counterpart \cite{Li_2019_CVPR} under this setting, which can be attributed two reasons: (1) we do not use the 2D-to-3D correspondences where the 3D poses are already in the camera frame, and (2) we add constraint to the camera estimation network based on a weak perspective camera assumption, which is not true for the Human3.6M dataset.

Following \cite{jahangiri2017generating, Li_2019_CVPR}, we also evaluate the robustness of the pose generator by testing on scenarios with missing joints. This is common in realistic scenarios when some joints are severely occluded and cannot be detected. During training, one or two missing joints are randomly selected from the the limb joints including l/r wrist, l/r knee, l/r elbow and l/r ankle. We use the ground truth 2D joints as input and set 2D coordinate of missing joints to zeros. The weights for different losses $\lambda_\text{gp}$, $\lambda_\text{2D}$, $\lambda_\text{reg}$ and $\lambda_\text{rec}$ are set to 0.1, 20.0, 7.5 and 10.0, respectively. We set the weights for missing joints in the 2D loss $\mathcal{L}_\text{2D}$ to zeros because missing joints do not provide any information for the training. The results are shown in Table \ref{Tab:missingJoints} where the numbers of \cite{Li_2019_CVPR,martinez2017simple, Wandt2019RepNet} are based on the public available implementation or checkpoints. We can see that our approach outperforms state-of-the-art weakly supervised approach\cite{Wandt2019RepNet}, and achieve comparable results with our supervised counterpart \cite{Li_2019_CVPR}.
\begin{table*}[h!]
\centering
\scriptsize
\setlength{\tabcolsep}{0.5pt}
\begin{tabular*}{0.99\textwidth}{ l c c c c c c c c c c c c c c c c c c } 
 \hline
 Algorithm & MH & WS &  Direct. & Discuss & Eating & Greet & Phone & Smoke & Pose & Purch. &  Sitting & SitD.  & Smoke & Wait & WalkD. & Walk & WalkT. & Avg.\\ 
 \hline

 Martinez \cite{martinez2017simple} &  & & 36.4 & 42.4 & 41.2 & 43.3 & 44.2 & 54.2 & 43.6 & 39.2 & 55.0 & 58.7 & 45.2 & 45.6 & 46.1 & 38.2 & 42.1 & 45.0
 \\
Jahangiri \cite{jahangiri2017generating} & \checkmark &  & 108.6 & 105.9 &105.6 &109.0 &105.5 & 109.9 &102.0 & 111.3 & 119.6 &107.8 & 107.1 &111.3 &108.4 & 107.0 & 110.3 & 108.6\\
Li \cite{Li_2019_CVPR} & \checkmark & & 31.4 & 38.5 & 37.1 & 37.8 & 40.2 & 49.0 & 37.1 & 35.1 & 47.8 & 56.7 & 40.7 & 39.5 & 40.9 & 31.2 & 34.7 & 39.8
 \\ 
  Wandt \cite{Wandt2019RepNet} & & \checkmark &36.9 & 42.2 & 36.5 & 43.7 & 41.4 & 46.7 & 40.4 & 42.0 & 48.7 & 57.3 & 42.0 & 43.4 & 42.9 & 38.4 & 38.4 & 42.7\\

 Ours &\checkmark & \checkmark & 35.4 & 41.3 & 33.7 & 42.3 & 39.1 & 47.1 & 36.2 & 46.9 & 46.4 & 57.7 & 38.6 & 43.0 & 42.0 & 34.8 & 37.0 & 41.2 \\
 
 \hdashline

 Martinez \cite{martinez2017simple} & & & 41.9 & 48.4 & 47.8 & 49.9 & 51.8 & 63.5 & 49.6 & 44.4 & 64.7 & 70.5 & 52.6 & 53.4 & 52.2 & 46.7 & 50.1 & 52.5
 \\
 
  Jahangiri \cite{jahangiri2017generating}  & \checkmark & & 125.0 &121.8 &115.1 &124.1 &116.9 &123.8 &116.4 &119.6 &130.8 & 120.6 & 118.4 & 127.1 &125.9 &121.6 &127.6 & 122.3\\
  
  Li \cite{Li_2019_CVPR} & \checkmark & & 36.7 & 42.4 & 41.6 & 43.6 & 46.6 & 57.0 & 42.7 & 39.9 & 57.0 & 65.8 & 46.8 & 45.4 & 46.5 & 36.3 & 41.0 & 46.0 
 \\  
  
 Wandt \cite{Wandt2019RepNet} & & \checkmark  & 52.2 & 62.2 & 48.4 & 59.5 & 56.7 & 70.6 & 53.9 & 57.8 & 61.5 & 83.5 & 57.7 & 58.6 & 73.9 & 58.2  & 62.8 & 60.8 \\

 Ours & \checkmark & \checkmark & 50.9 & 53.9 & 49.8 & 54.8 & 54.7 & 65.1 & 49.4 & 49.3 & 63.5 & 76.1 & 54.5 & 54.3 & 59.8 & 54.8 & 56.1 & 56.4 
 \\
 \hline
\end{tabular*}
\caption{Results with one (the first five rows) or two (the last five rows) missing joints.}
\label{Tab:missingJoints}
\vspace{-3mm}
\end{table*}

\subsection{Ablation Studies}

\begin{figure*}[h!]
\begin{center}
\includegraphics[width=0.98\linewidth]{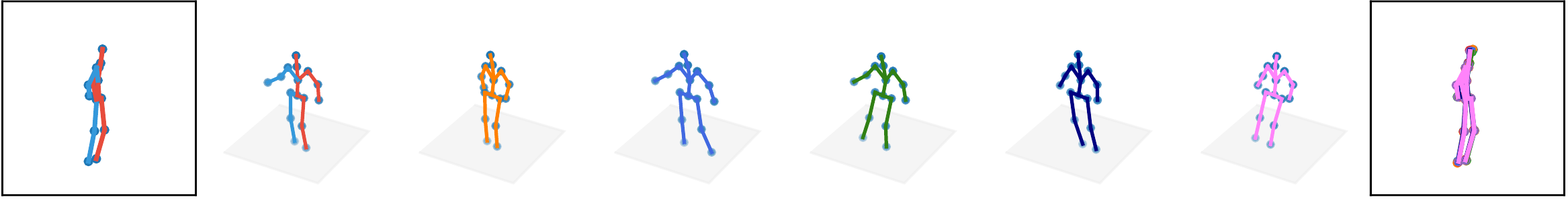}\\
\includegraphics[width=0.98\linewidth]{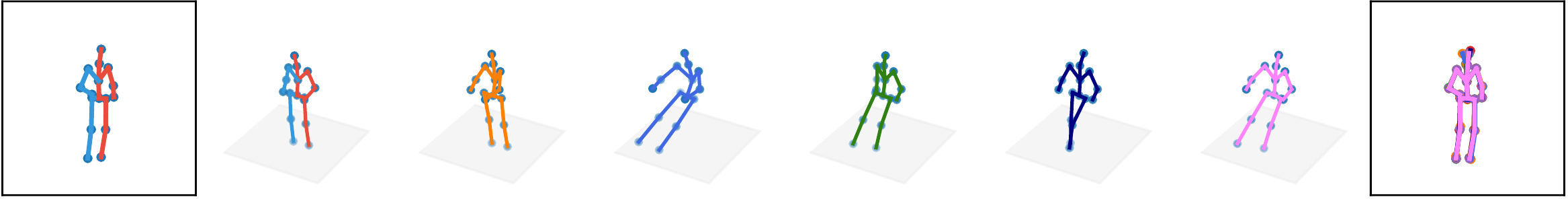}\\
\includegraphics[width=0.98\linewidth]{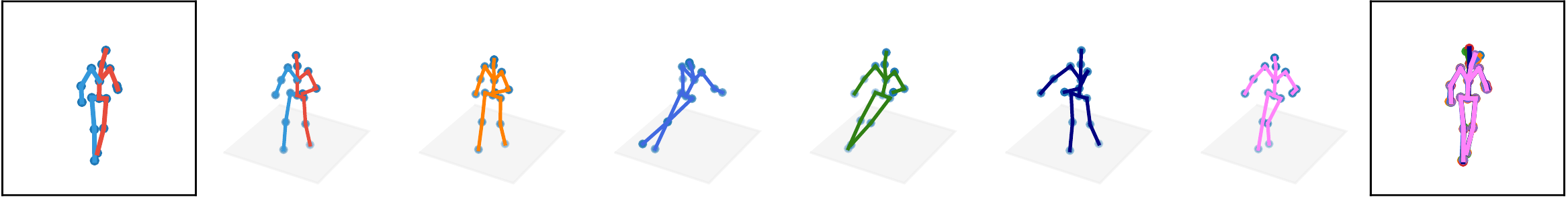}\\

\end{center}
\vspace{-6mm}
   \caption{Visualization of five hypotheses sampled by FPS (third to seventh columns). The first and second columns represent the input 2D pose and the corresponding 3D ground truth. The last column shows the 2D reprojections of the five hypotheses (the corresponding 2D reprojection and 3D pose are drawn in the same color).}
\label{fig:multi-hypotheses}
\end{figure*}

\paragraph{Do $\mathcal{L}_\text{reg}$ and $\mathcal{L}_\text{rec}$ prevent model collapse?}We compare our model with and without $\mathcal{L}_\text{reg}$ and $\mathcal{L}_\text{rec}$ to verify their effectiveness on diversity. 
We do the evaluation on two metrics: (1) randomly sample 10 pose hypotheses for the same 2D input and calculate the standard deviation (STD) of each joint coordinate \wrt the root joint;
(2) use the farthest point sampling (FPS) \cite{eldar1997farthest} to sample 5 diverse hypotheses from 100 random samples and compute the standard deviation (STD-FPS).  
Table~\ref{diversity} shows the MPJEP under best hypothesis(BH) and zero code(MS) settings, STD and STD-FPS of our model with and without $\mathcal{L}_\text{reg}$ and $\mathcal{L}_\text{rec}$. We can see that the pose hypotheses generated by our full model is much more diverse than the model without $\mathcal{L}_\text{reg}$ and $\mathcal{L}_\text{rec}$. Moreover, the full model achieves lower error shows the advantage of generating diverse hypotheses. We also show the five hypotheses sampled by FPS in Figure~\ref{fig:multi-hypotheses}. We can see that the generated poses have different degree of diversity depending on the input 2D poses. The 2D reprojections of all 3D pose hypotheses (last column) overlap with each other shows that there are multiple solutions for each 2D input.

\begin{table*}[ht]
\vspace{-3mm}
\centering
\subfloat[]{
\scriptsize
\setlength{\tabcolsep}{3pt}
\begin{tabular}[t]{ c c c c c } 
 \hline
 Model  & MPJPE(BH) & MPJPE(ZC) & STD & STD-FPS\\ 
 \hline
  Full model    &31.6 &35.3   & 77.4 & 122.3 \\
  \hline
  Without     &36.3  &37.4  & 3.6 & 4.8 \\ 
  \hline
\end{tabular}
\label{diversity}
}\quad \quad
\subfloat[]{
\scriptsize
\setlength{\tabcolsep}{3pt}
\begin{tabular}[t]{ c c c c c c c c} 
 \hline
 $\lambda_{\text{reg}}$ &  7.0 & 7.5 & 8.0 & 9.0 & 10.0 & 11.0 & 12.0\\ 
 \hline
  STD     & 72.0   & 77.4 & 81.0 & 91.5 & 98.2 & 108.2 & 113.7\\
  \hline
  MPJEP   &33.0  & 31.6  & 31.9 & 32.8 & 34.2 & 36.1 & 38.3\\ 
  \hline
\end{tabular}
\label{accuracy and diversity}
}
\caption{(a):Our model with and without $\mathcal{L}_\text{reg}$ and $\mathcal{L}_\text{rec}$. (b): The impact of changing the weights $\lambda_{\text{reg}}$ on the diversity and accuracy}
\vspace{-5mm}
\end{table*}

\paragraph{How does $\lambda_{\text{reg}}$ affect the diversity and accuracy?} We add the $\mathcal{L}_\text{reg}$ to explicitly encourage the diversity of the generated 3D poses, here we analyze the impact of
changing the corresponding weight $\lambda_{\text{reg}}$. Table \ref{accuracy and diversity} shows the estimation error under BH setting and diversity (STD) when $\lambda_{\text{reg}}$ is set to 7.0, 7.5, 8.0, 9.0, 10.0, 11.0, 12.0 with weights for other losses fixed. We can see that the STD increases when  $\lambda_{\text{reg}}$ gets larger, which verifies that the $\mathcal{L}_\text{reg}$ helps to increase the diversity. At the same time, the error also becomes large where we impose overly strong constraint on diversity with high $\lambda_{\text{reg}}$. Consequently, the value of $\lambda_{\text{reg}}$ should be a trade-off between accuracy and diversity.


 



\vspace{-3mm}

\vspace{-4mm}
\begin{figure*}[h!]
\centering
\includegraphics[width=0.23\textwidth]{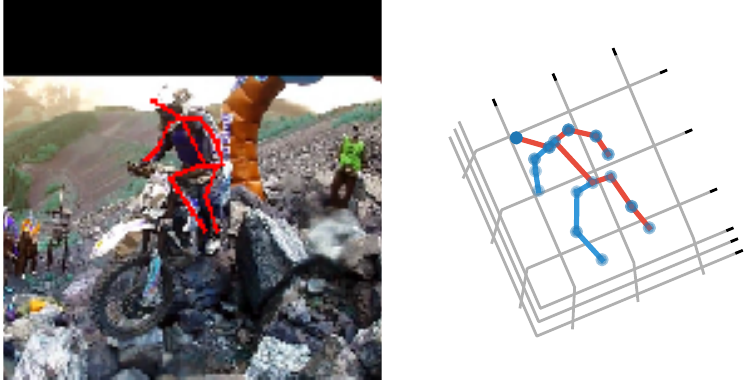} 
\includegraphics[width=0.23\textwidth]{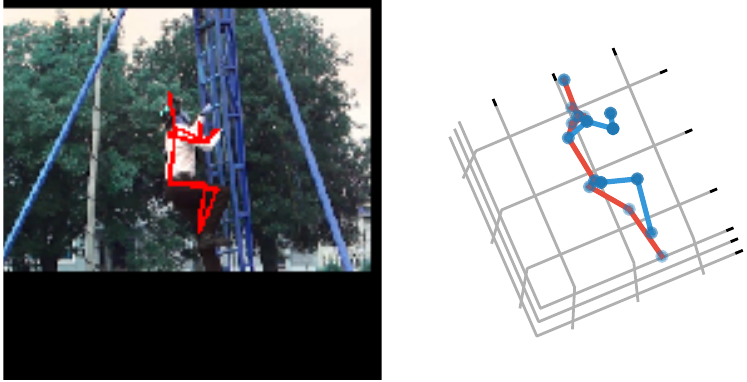}
\includegraphics[width=0.23\textwidth]{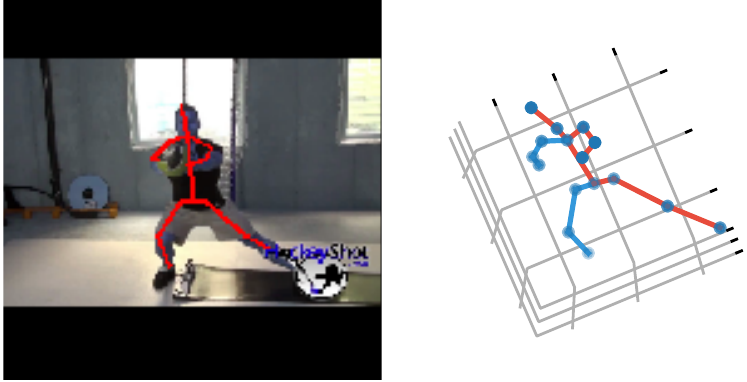} 
\includegraphics[width=0.23\textwidth]{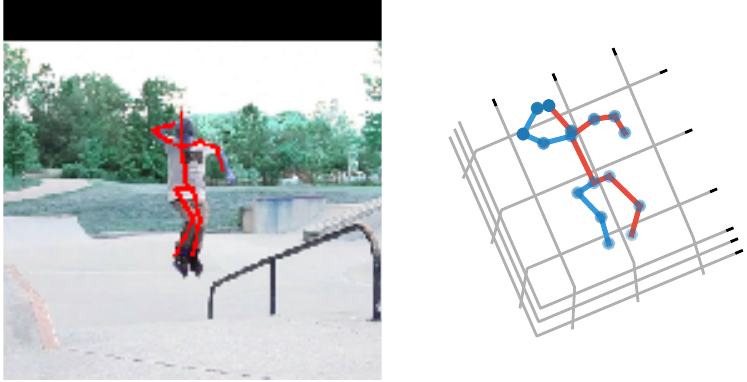} \\
\includegraphics[width=0.23\textwidth]{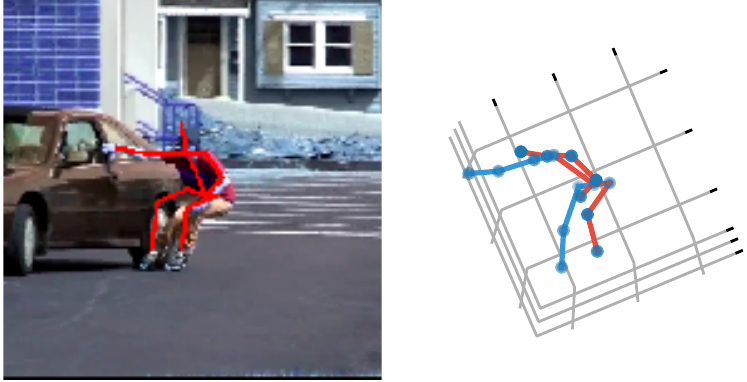} 
\includegraphics[width=0.23\textwidth]{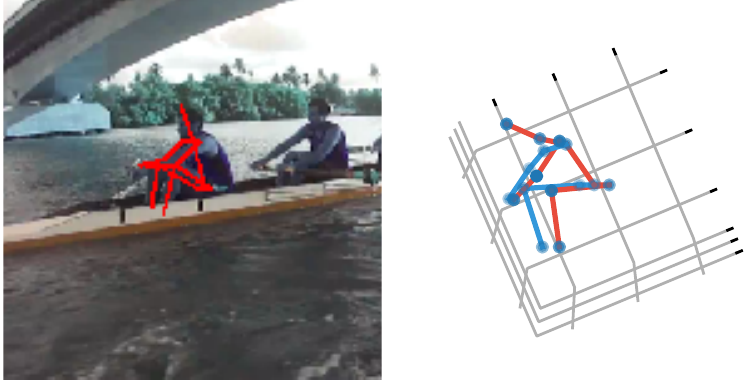} 
\includegraphics[width=0.23\textwidth]{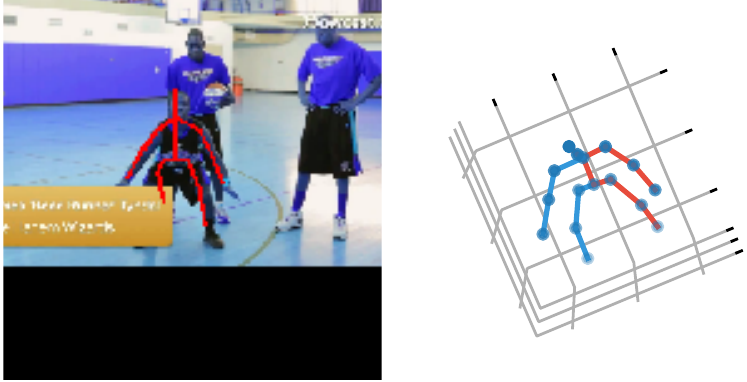} 
\includegraphics[width=0.23\textwidth]{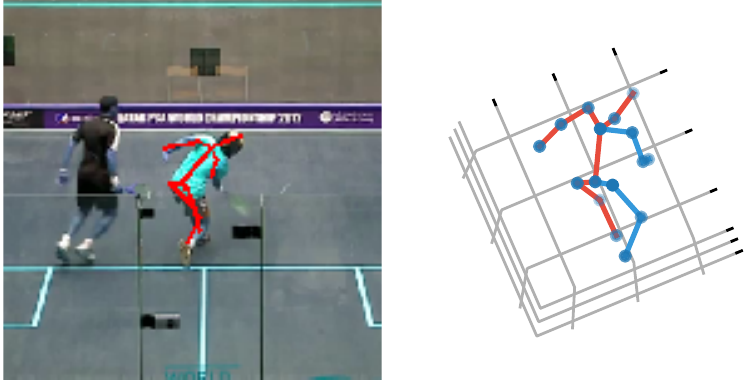} \\
\includegraphics[width=0.23\textwidth]{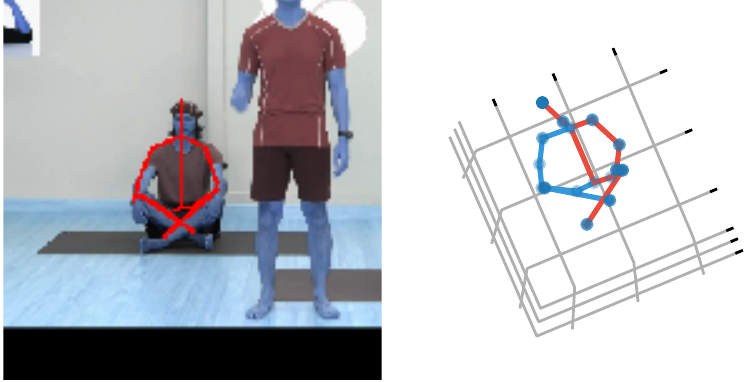} 
\includegraphics[width=0.23\textwidth]{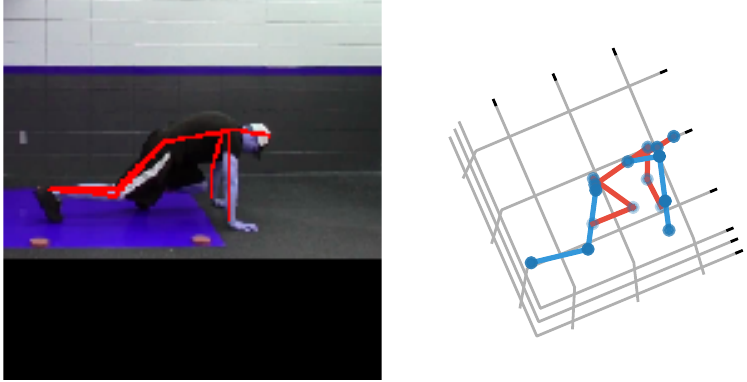}
\includegraphics[width=0.23\textwidth]{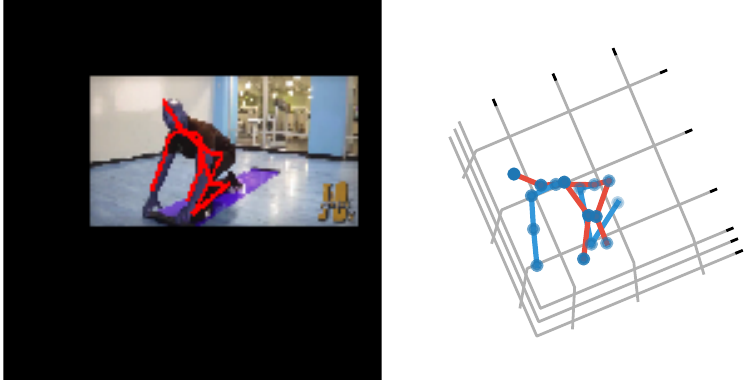} 
\includegraphics[width=0.23\textwidth]{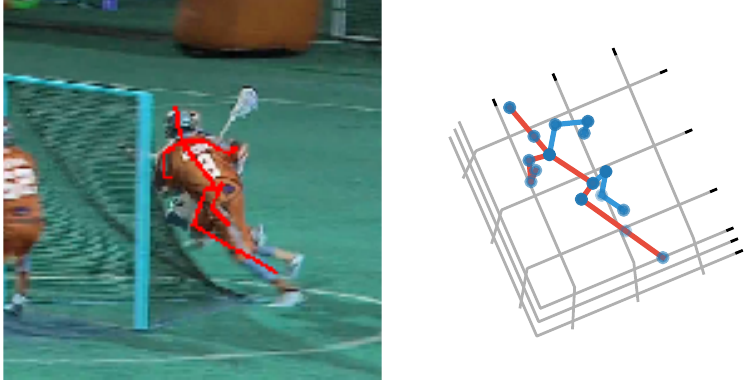} \\
\vspace{-3mm}
\caption{Qualitative results on the MPII dataset.}
\label{fig:viz_mpii}
\vspace{-5mm}
\end{figure*}

\begin{wraptable}{r}{7cm}
\vspace{-4mm}
\centering
\scriptsize
\setlength{\tabcolsep}{2.5pt}
\begin{tabular*}{0.48\textwidth}{ l c c c c c c } 
 \hline

 Algorithm  & MH & WS & GS & No GS & Outdoor & All PCK\\ 
 \hline

 Mehta* \cite{mehta2017monocular} & & & 84.1 & 68.9 & 59.6  &  72.5  \\
 Li \cite{Li_2019_CVPR} &\checkmark & & 70.1 & 68.2 & 66.6 & 67.9\\
 Kanazawa \cite{kanazawa2018end} & & \checkmark & - & - & - & 77.1  \\
 Wandt \cite{Wandt2019RepNet} & & \checkmark & - & - & - & 81.8  \\
 Ours(ZC) & \checkmark &\checkmark  &82.1 & 81.0 & 72.3  & 79.3 \\ 
 Ours(BH) & \checkmark &\checkmark  &86.9 & 86.6 & 79.3  & 85.0 \\ 
 \hline

  \hline
\end{tabular*}
\caption{Results on the MPI-INF-3DHP dataset.}
\label{Tab:resultsMPI-INF-3DHP}
\end{wraptable}

\vspace{5mm}
\subsection{Results on MPI-INF-3DHP and MPII datasets}
We test the generalization capacity of our approach on the MPI-INF-3DHP and MPII datasets. The MPI-INF-3DHP dataset includes images under three different scenes: indoor images with (GS) and without green screen background (no GS), outdoor images (Outdoor), and the MPII dataset only includes outdoor images. Table~\ref{Tab:resultsMPI-INF-3DHP} shows the quantitative results of our approach under ZC and BH settings for the MPI-INF-3DHP dataset. Our results is slightly worse than \cite{Wandt2019RepNet} under ZC setting but outperforms other approaches under BH setting.  We only show qualitative results for the MPII dataset because the 3D ground truth is not available. As can been seen from Figure~\ref{fig:viz_mpii}, our approach generalizes well to outdoor scenes.

\section{Conclusion}

We propose a weakly supervised generative network for 3D human pose estimation. Our network is designed to model a proposal distribution and learned by minimizing the KL divergence with the true posterior distribution. Experiments show that our network is able to generate feasible 3D pose hypotheses consistent with 2D reprojections and also achieves better results compared to existing weakly supervised approaches. Moreover, results on the MPII and MPI-INF-3DHP datasets verify the generalization capacity of our network.

\bibliography{egbib}
\end{document}